\documentclass[10pt,twocolumn,letterpaper]{article}

\usepackage{cvpr}
\usepackage{times}
\usepackage{epsfig}
\usepackage{graphicx}
\usepackage{amsmath}
\usepackage{amssymb}
\usepackage{booktabs}
\usepackage{makecell}
\usepackage{dsfont}

\usepackage[breaklinks=true,bookmarks=false]{hyperref}

\cvprfinalcopy 

\def\cvprPaperID{****} 

\begin{document}

\title{Distance Guided Channel Weighting for Semantic Segmentation}

\author{Xuanyi Liu$^{1}$ \quad Lanyun Zhu$^{2}$ \quad Shiping Zhu$^{2}\thanks{Corresponding author.}$ \quad Li Luo$^{2}$\\
{$^{1}$Nanyang Technological University} \quad {$^{2}$Beihang University}\\
{\tt\small \{zhulanyun, shiping.zhu\}@buaa.edu.cn}\quad{\tt\small {lxycopper@gmail.com}\quad{\tt\small louieluo@buaa.edu.cn}}
}

\maketitle


\begin{abstract}
    Recent works have achieved great success in improving the performance of multiple computer vision tasks by capturing features with a high channel number utilizing deep neural networks. However, many channels of extracted features are not discriminative and contain a lot of redundant information. In this paper, we address above issue by introducing the Distance Guided Channel Weighting (DGCW) Module. The DGCW module is constructed in a pixel-wise context extraction manner, which enhances the discriminativeness of features by weighting different channels of each pixel’s feature vector when modeling its relationship with other pixels. It can make full use of the high-discriminative information while ignore the low-discriminative information containing in feature maps, as well as capture the long-range dependencies. Furthermore, by incorporating the DGCW module with a baseline segmentation network, we propose the Distance Guided Channel Weighting Network (DGCWNet). We conduct extensive experiments to demonstrate the effectiveness of DGCWNet. In particular, it achieves 81.6\% mIoU on Cityscapes with only fine annotated data for training, and also gains satisfactory performance on another two semantic segmentation datasets, i.e. Pascal Context and ADE20K. Code will be available soon at \url{https://github.com/LanyunZhu/DGCWNet}.
\end{abstract}

\section{Introduction}
%
Semantic segmentation aims to assign a semantic label for each pixel in an image. It is broadly applied in a lot of areas such as automatic driving and robot sensing. Machine learning based methods treat semantic segmentation as a pattern recognition problem, and tackle it by classifying each pixel utilizing the extracted features. Traditional methods usually extract features manually, which lacks robustness, thus failing to achieve satisfactory performance under complex conditions. Recently, benefited from the rapid development of deep learning, many CNN-based methods \cite{fcn, deeplabv3, pspnet} have achieved great success. By automatically learning rather than manually extracting useful features, deep neural networks are able to capture the more informative and discriminative representations. 
\begin{figure}[t]
    \centering
    \includegraphics[width=2.8in]{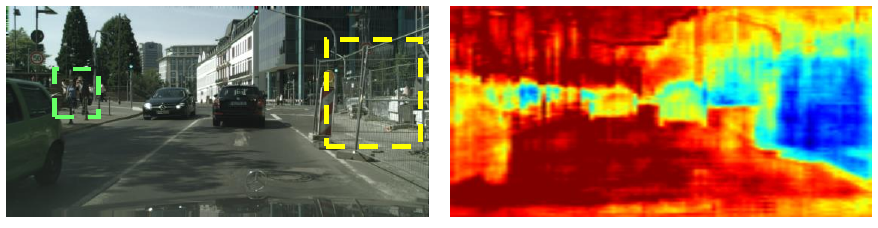}
    \caption{The visualization of a channel's map from the last layer of Deeplabv3. Two regions marked by green box and yellow box respectively, belong to different categories, while have similar values on this channel.}
    \label{fig1}
\end{figure}

Many CNN-based methods \cite{vgg, resnet} enhance the representativeness of models by extracting features with more channels, which can help to encode richer properties. It is expected that every channel can represent certain discriminative properties behaving differently on different classes, thus benefiting to the distinction of different categories. However, we find some channels in existing CNN models lack the inter-class diversity. For example, as shown in Fig.~\ref{fig1}, we visualize a channel $c$ of feature maps from the last layer of Deeplabv3. It can be observed that, though regions marked by two boxes belong to different categories, the values of this channel on them are very similar. We call properties represented by this kind of channels as low-discriminative redundant properties, 
~as the similar performance on different categories adversely affect the inter-class distinction. It is obvious that the low-discriminative properties are useless and even harmful for the semantic prediction.

To address the above issue, traditional machine learning methods always employ dimensionality reduction approaches, such as Principal Component Analysis (PCA) and Independent Component Correlation (ICA). The essence of dimensionality reduction lies in finding a subset of input feature or reconstructing a set of new features that are more discriminative than the original one. In the task of semantic segmentation, the new features are able to represent the most crucial semantic concepts. While performance of traditional machine learning methods have been improved by deep CNN greatly, the problem of redundant high-dimensional features in CNN model leads us to think: can we leverage the idea of dimensionality reduction with the powerful feature extraction ability of deep learning?  If we can design an approach to fully utilize those high-discriminative properties while ignore the low-discriminative properties, the semantic segmentation will become easier.

Motivated by the above discussion, in this paper, we propose a novel Distance Guided Channel Weighting (DGCW) Module, a simple unit to focus on the high-discriminative properties as well as capture the long-range pixel-wise relationship. The function principle of enhancing the representativeness of features is motivated by dimensionality reduction approaches such as PCA and ICA. However, different from PCA and ICA that reconstruct features guided by global statistics such as variance, DGCW achieves it from a pixel-wise perspective, i.e. modeling the relationship between every two pixels. Specifically, for each pixel in the feature map, DGCW correlates it with all other pixels in a self-adaptive manner, then aggregates all correlation features for feature updating. The correlation process is simple yet explicit. That is, when correlating pixel $i$ and $j$, DGCW selectively emphasizes and de-emphasizes different channels of $i$'s feature vector through channel-wise weighting, where the weight values are obtained from the channel-wise distance of $i$ and $j$. More specifically, the higher-distance channels, representing more discriminative properties for $i$ and $j$, obtain larger weight values to be emphasized. The pixel-wise perspective setting is inspired by the success of various global context extraction approaches \cite{pspnet, bisenet}, especially the self-attention mechanism represented by non-local networks \cite{ocnet, ccnet, ann}. Under this setting, the feature from each pixel can perceive features of all other pixels. Thus, the pixel-wise long-range dependencies are exploited. 

We further embed the proposed DGCW module into a base network, getting a new model for semantic segmentation, called as DGCWNet. We evaluate our methods on three challenging datasets, including Cityscapes, PASCAL Context and ADE20K. DGCWNet achieves top performance on all three datasets, gaining 81.16\%, 53.9\% and 45.51\% mIoU, respectively. We also conduct extensive experiments with various base networks, demonstrating that DGCW module can improve the performance effectively when combining with different existing networks for semantic segmentation.  
\section{Related Work}
\noindent \textbf{Semantic Segmentation.} In recent years, semantic segmentation has achieved great success with the rapid development of deep neural networks. FCN \cite{fcn} first replaces the fully connected layers in common classification networks by convolution layers to generate segmentation masks. However, the high-scale downsampling leads to the loss of spatial details. Some works \cite{unet, refinenet, segnet, dfn} address this limitation by employing an encoder-decoder structure to fuse low-level spatial information and high-level semantic information. Deeplabs \cite{deeplabv1, deeplabv2, deeplabv3, deeplabv3+} utilize atrous convolution to extract features from a large receptive field while maintaining feature map's size. Some works are focused on capturing and aggregating multi-level contextual information. PSPNet \cite{pspnet} performs pooling operation at different grids to extract and fuse features from different levels. Deeplabv3 \cite{deeplabv3} employs a pyramid structure composed of atrous convolutions with multiple atrous rates, called as ASPP. DenseASPP \cite{denseaspp} further brings dense connections into ASPP to extract richer and more complex features.
\\ \hspace*{\fill} \\
\begin{figure*}[t]
    \centering
    \includegraphics[width=7in]{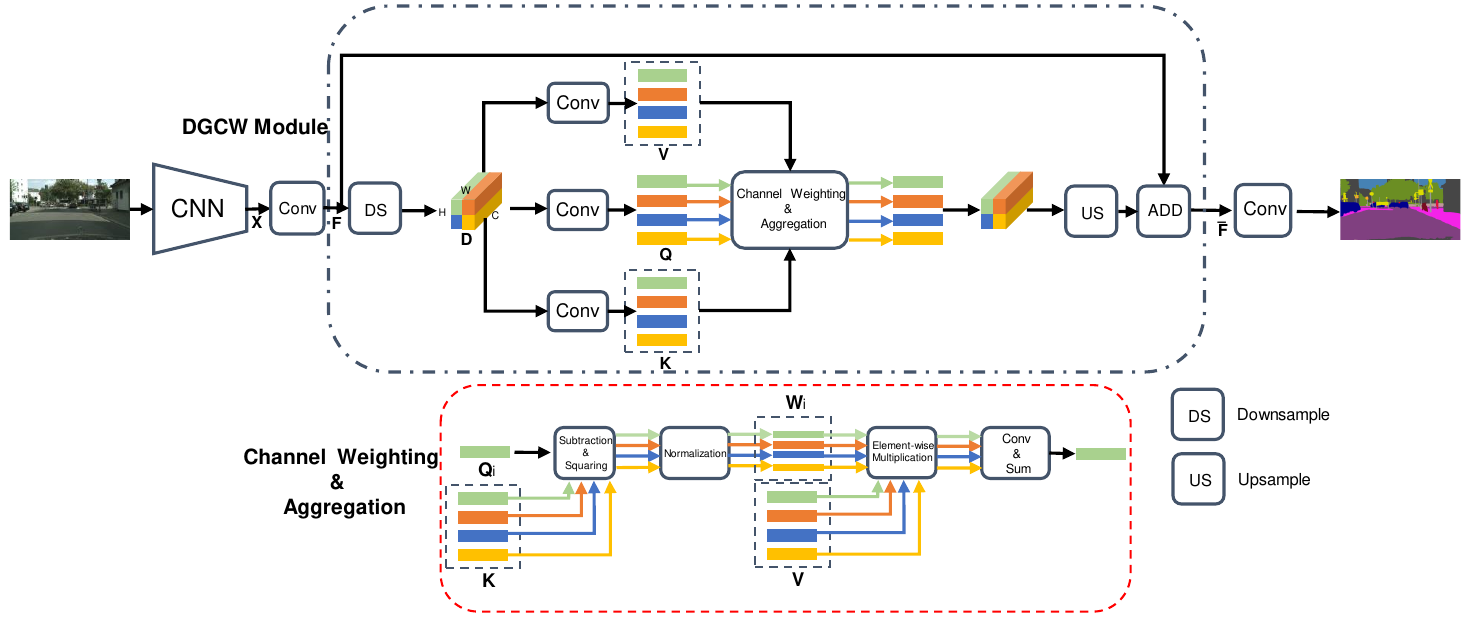}
    \caption{The overall structure of DGCWNet. For easier description, DGCW module’s input feature map shown in this picture has a resolution of $2\times 2$.}
    \label{overall}
\end{figure*}
\noindent \textbf{Image-level global context.} Global context has shown its effectiveness in improving the performance of semantic segmentation tasks. Some works directly utilize global average pooling to aggregate features from all pixels, thus generating image-level global information representation. BiseNet \cite{bisenet} adds the global pooling at the end of context path and DFN \cite{dfn} adds it on the top of base network. Moreover, inspired by the squeeze-and-excitation attention mechanism \cite{senet}, \cite{bisenet, dfn} further extract channel-wise relationship utilizing features captured by global average pooling. However, the global pooling treats all pixels equally rather than distinguishing pixels from different classes when exploiting the global context. To address the issue, a lot of methods are focused on generating the pixel-wise long-range contextual information. Some works employ the self-attention mechanism to aggregate features of pixels that belong to the same category. OCNet \cite{ocnet} updates feature vector of each pixel by weighting and summing features from all pixels, and the weight value is obtained from the inter-pixel similarity. DANet \cite{danet} performs both channel-wise self-attention and spatial-wise self-attention. CCNet \cite{ccnet} reduces the huge computation cost of traditional self-attention modules by employing a criss-cross attention. ACFNet \cite{acfnet} calculates the class center representing the average feature of all pixels belonging to each class to exploit the categorical context. Our DGCWNet is different from the aforementioned methods that correlate all pixels by aggregating features from different pixels. In contrast, DGCWNet models the relationship between every two pixels by emphasizing and de-emphasizing different dimensions of every pixel's feature vector guided by the channel-wise distance. 
\noindent\section{Method}
\subsection{Overall}
The overall network structure is shown in Fig.~\ref{overall}. The input image is first fed into a fully-convolutional network, yielding a high-dimensional feature map $\textbf{X}$. After that, to reduce the computation cost and memory usage, we apply a convolution layer on $\textbf{X}$ for channel reduction, obtaining a low-dimensional feature map $\textbf{F}$ with a channel number of 128. Then, $\textbf{F}$ is fed into the proposed DGCW module, outputting an updated feature map $\Bar{\textbf{F}}$. Finally, a classifier convolution is followed up to predict the label for each pixel. The whole network is called DGCWNet. 

\subsection{DGCW Module}
We propose a DGCW module to model the long-range contextual dependencies over different pixels for an image. Consider the input feature map $\textbf{F}$, to reduce computation, we first downsample it with a ratio of 4, getting  $\textbf{D}\in \mathbb{R}^{C\times H\times W}$, where $C$, $H$ and $W$ denote feature channel number (a.k.a. dimensional number), height and width, respectively. Then we flat $\textbf{D}$ to size $\mathbb{R}^{C\times HW}$. After that, we apply three $1\times 1$ convolution layers $W_{q}$, $W_{k}$ and $W_{v}$ on $\mathbf{F}$ to generate three feature maps $\mathbf{Q}$, $\mathbf{K}$ and $\mathbf{V}$:
\begin{equation}
    {\mathbf Q} = W_{q}({\mathbf D}); {\mathbf K} = W_{k}({\mathbf D}); {\mathbf V} = W_{v}({\mathbf D}),
\end{equation}
At each row $i$ of feature matrix $\mathbf{Q}$, we can get a feature vector $\mathbf{Q}_{i}\in \mathbb{R}^{C}$. Similarly, we can obtain a vector $\mathbf{K}_{j}\in \mathbb{R}^{C}$ at each row $j$ of matrix $\mathbf{K}$. We further generate the channel-wise distance maps by subtracting and squaring each $\mathbf{Q}_{i}$ and each $\mathbf{K}_{j}$:
\begin{equation}
    \mathbf{M}_{i,j} = (\mathbf{Q}_{i} - \mathbf{K}_{j})^{2},
\end{equation}
where $\mathbf{M}_{i,j}\in \mathbb{R}^{C}$ denotes the channel-wise distance between $\mathbf{Q}_{i}$ and $\mathbf{K}_{j}$, ${i,j}\in [0, ..., HW-1]$. After that, we employ a normalization $f$ on $\mathbf{M}$ to obtain the channel-wise weight values map $\mathbf{W}\in \mathbb{R}^{C\times HW\times HW}$:
\begin{equation}\label{get_weight}
    \mathbf{W} = f(\mathbf{M}),
\end{equation}
The normalization layer can be chosen flexibly, and we find that simply dividing by the sum of the vector's all channels can yield the best performance. We will show latter in subsection \ref{norm} the performance comparison of employing different normalization layers. In this setting, weight values of low-distance channels are smaller than high-distance channels. Extremely, if $\mathbf{Q}_{i}$ and $\mathbf{K}_{j}$ are exactly equal on a channel $c$, the corresponding weight value $\mathbf{W}_{c,i,j}$ will be 0. 

Then we utilize $\mathbf{W}$ to weight different channels of $\mathbf{V}$ through element-wise multiplication, and feed the weighted feature map into two convolution layers, getting the pixel-wise relationship map $\mathbf{R}\in \mathbb{R}^{C\times HW\times HW}$.
Specifically, each element $\mathbf{R}_{i,j}\in \mathbb{R}^{C}$ of $\mathbf{R}$ represents the relationship vector of feature vectors from $i$ and $j$, which can be calculated as:
\begin{equation}
    \mathbf{R}_{i,j} = g(\mathbf{W}_{i,j}\cdot\mathbf{V}_{i}),
\end{equation}
where feature vector $\mathbf{V}_{i}$ is the $i'{th}$ row of $\mathbf{V}$, $g$ denotes two convolution layers, in which the first one is followed by a ReLU to increase nonlinearity. Finally, we aggregate all relationship features for each pixel by performing a sum operation for $\mathbf{R}$ on the third dimension, then upsample with a ratio of 4, and add it with input $\mathbf{F}$ to get the updated feature map $\Bar{\mathbf{F}}$:
\begin{equation}
    \Bar{\mathbf{F}_{i}} = \mathbf{F}_{i} + \rm{US}\left(\sum_{j\in [0,...,C-1]}\mathbf{R}_{i,j}\right).
\end{equation}
where $\Bar{\mathbf{F}_{i}}$ denotes a feature vector of updated $\Bar{\mathbf{F}}$ at pixel $i$, $\rm{US}$ refers to upsampling operation.

The benefits of DGCW module are two-fold. First, it enables the feature of each pixel to perceive the features of all other pixels, thus generating more powerful pixel-wise global representation. Moreover, it forces network to pay more attention to the high-discrminative properties when correlating every two pixels. For example, if a channel represent properties that are highly correlated to the prediction of class $i$ while behave similarly on other categories, the DGCW module will ignore these properties when modeling the relationship between two pixels belonging to class $j$ and $k$.

\subsection{Relationship with Self-attention}
Self attention mechanism, represented by non-local mechanism, is also able to capture the long-range pixel-wise relationship. The output $\Bar{\mathbf{F}}$ of a self-attention module is computed as:
\begin{equation}
\Bar{\mathbf{F}} = \mathbf{F} + h\left(g\left(f\left(W_{1}\left(\mathbf{F}\right), W_{2}\left(\mathbf{F}\right)\right), W_{3}\left(\mathbf{F}\right)\right)\right).
\end{equation}
where $\mathbf{F}$ is the input feature map, $W_{1}$, $W_{2}$, $W_{3}$, $h$ represent four convolution layers, $f$ represents matrix multiplication followed by a softmax function, $g$ represents another matrix multiplication. 

The operation process of our proposed DGCW module can also be described by this formula. The difference is, in DGCW module, $f$ represents the pixel-wise vectors subtraction and squaring followed by a normalization layer, $g$ represents element-wise multiplication. 

The self-attention module and DGCW module share the same general formula and both of them can model the inter-pixel relationship to extract global context. Thus, DGCW module can be treated as a special form of self-attention. However, the key difference lies in the way correlating features of every two pixels. 
Specifically, self-attention module extracts pixel-wise global context by aggregating features of all pixels weighted by the intra-class similarity. Essentially, every pixel of the output feature map is a linear combination of feature vectors from all pixels of the input map. 
Thus, every two pixels $i$ and $j$ are correlated by weighting $i$'s feature vector $f_{i}$ and adding it to output's pixel $j$.
~In contrast, DGDW module correlates $i$ and $j$ by emphasizing the high-distance channels and ignoring low-distance channels between feature vectors $f_{i}$ and $f_{j}$ of $i$ and $j$, respectively. 

\section{Experiments}
\subsection{Network Structure}
We apply three base networks to demonstrate the generality and effectiveness of proposed DGCW module. The first one is ResNet-101, which is the default backbone of all networks in experiments. The other are ResNet-101 with Pyramid Pooling Module (PPM) \cite{pspnet} and ResNet-101 with Atrous Sptial Pyramid Pooling (ASPP) \cite{deeplabv3}. We conduct experiments on different base networks to verify that our proposed method has strong generality. 

\textbf{ResNet-101.} We use ResNet-101 pretarined on ImageNet as the backbone of our network. Following some previous works \cite{pspnet, deeplabv3}, we remove the last two down-sampling operations and apply the dilated convolution instead, thus holding output feature map $\frac{1}{8}$ of the input image. 

\textbf{ResNet-101 with PPM.} Pyramid Pooling Module (PPM) is an effective way to encode multi-level global features. Following the original paper, we perform pooling operation at different grid scales, including $1\times 1$ region, $2\times 2$ regions, $3\times 3$ regions and $6\times 6$ regions. After that, we feed four feature maps with different scales into $1\times 1$ convolution layer for updating, then upsample them to the same size. Finally, we connect four updated feature maps and the input feature map on the channel dimension to get the output of PPM.  

\textbf{ResNet-101 with ASPP. }The Atrous Spatial Pyramid Pooling (ASPP) module is composed of five parallel branches, including a global average pooling branch, a $1 \times 1$ convolution branch and three $3\times 3$ convolution branches with different dilation rates being 12, 24 and 36, respectively. For a fair comparison, following \cite{ocnet, acfnet}, we change the output channel number from 256 to 512 for all five parallel branches. 

\subsection{Datasets}
We evaluate our method on three popular datasets, including Cityscapes, PASCAL Context and ADE20K. 

\textbf{Cityscapes.} Cityscapes is a large urban street scene parsing dataset. It contains 5000 fine annotated images and 20000 coarse annotated images. We only utilize the fine annotated images for experiments. The fine annotated images are divided into 2975 for training, 500 for validation and 1525 for testing. The resolution of each image is $2048\times 1024$ and all pixels are labeled to 30 classes. Following \cite{pspnet}, we only use 19 most common classes for training and testing. 

\textbf{PASCAL Context.} PASCAL Context provides the detailed segmentation labels for images of PASCAL VOC 2010. It contains 4998 images for trainng and 5105 images for validation. Following \cite{danet}, we use the most frequent 59 classes along with a background class for training and evaluating. 

\textbf{ADE20K.} ADE20K is a large and high-quality semantic segmentation benchmark with 150 object and stuff categories. The dataset is divided into 20K/2K/3K images
for training, validation, and testing, respectively.

\subsection{Implementation Details}
\textbf{Loss Setteings. }Following \cite{pspnet, ccnet}, to make the deep network easier to train, we utilize the auxiliary loss in all experiments. In addition to supervising the final output of DGCWNet, we also supervise the output of backbone ResNet Stage4. Both loss functions are cross entropy loss. The overall loss can be denoted as:
\begin{equation}
    \mathcal{L} = \mathcal{L}_{f} + \lambda \mathcal{L}_{4}.
\end{equation}
where $\mathcal{L}_{f}$ refers to loss of final output and $\mathcal{L}_{4}$ is loss of ResNet Stage4's output. $\lambda$ is set to 0.4. 
\begin{table}[t]
    \centering
    \begin{tabular}{l c}
       \toprule
       Method & Mean IoU(\%)\\
       \midrule
       ResNet-101  &  75.65\\
       ResNet-101 + DGCW & 77.81\\
       ResNet-101 + conv & 75.90\\
       ResNet-101 + DGCW(softmax) & 77.46\\
       ResNet-101 + DGCW(tanh) & 77.44\\
       \midrule
       ResNet-101-PPM &  77.88\\
       ResNet-101-PPM + DGDW & 79.60\\
       ResNet-101-PPM + conv & 78.19\\
       ResNet-101-PPM + DGCW(softmax) & 79.19\\
       ResNet-101-PPM + DGCW(tanh) & 79.14\\
       \midrule
       ResNet-101-ASPP &  78.61\\
       ResNet-101-ASPP + DGCW & 80.43\\
       ResNet-101-ASPP + conv & 78.75\\
       ResNet-101-ASPP + DGCW(softmax) & 79.95\\
       ResNet-101-ASPP + DGCW(tanh) & 79.92\\
       \bottomrule
    \end{tabular}
    \caption{Ablation studies of DGCW module on Cityscapes validation set.}
    \label{ablation1}
\end{table}

\textbf{Training Settings. }We apply the Stochastic Gradient Descent (SGD) as the optimizer of all experiments, in which we set the weight decay to 0.0005 and momentum to 0.9. We use the `poly' policy to set the learning rate during the training process, where the learning rate of each iteration equals to the initial rate multiplied by $(1-\frac{iter}{max_{iter}})^{0.9}$. We set initial learning rate to 0.01 for Cityscapes, 0.005 for PASCAL Context and 0.02 for ADE20K. Following \cite{ccnet}, we adopt some widely used data augmentation strategies to avoid overfitting, including random cropping, random flipping and random scaling in the range of 0.5 to 2.0. We set the crop size to $769\times 769$ for Cityscapes and $513\times 513$ for PASCAL Context and ADE20K. All Batchnorm layers in the networks are replaced by Sync-BN and the batch size is 8 for experiments on Cityscapes and 16 for other datasets. We employ 8 $\times$ NVIDIA 2080Ti GPUs to conduct experiments. 

\subsection{Experiments on Cityscapes}
\subsubsection{Ablation Studies} \label{norm}
To demonstrate the effectiveness of proposed DGCWNet, we conduct some ablation experiments on Cityscapes validation set.\\
\begin{table}[t]
    \centering
    \begin{tabular}{l c}
        \toprule
        Method & Mean Iou(\%)\\
        \midrule
        ResNet-101 & 75.65\\
        ResNet-101 + DGCW & 77.81\\
        ResNet-101 + GAP & 76.01\\
        ResBet-101 + SE & 76.22\\
        ResNet-101 + NL-H & 77.34\\
        ResNet-101 + NL-D & 76.85\\
        \midrule
        ResNet-101-PPM & 77.88\\
        ResNet-101-PPM + DGCW &79.60\\
        ResNet-101-PPM + GAP & 78.25\\
        ResNet-101-PPM + SE & 78.41\\
        ResNet-101-PPM + NL-H & 78.79\\
        ResNet-101-PPM + NL-D & 78.03\\
        \midrule
        ResNet-101-ASPP & 78.61\\
        ResNet-101-ASPP + DGCW & 80.43\\
        ResNet-101-ASPP + GAP & 78.78\\
        ResNet-101-ASPP + SE & 78.83\\
        ResNet-101-ASPP + NL-H & 79.60\\
        ResNet-101-ASPP + NL-D & 78.97\\
        \bottomrule
    \end{tabular}
    \caption{Ablation studies of different global context extraction methods on Cityscapes validation dataset.}
    \label{ablation2}
\end{table}
\noindent \textbf{Ablation of DGCW Module. }We first verify the effect of DGCW module. Experimental results are shown in Tab.~\ref{ablation1}. We conduct experiments on three base networks, including ResNet-101, ResNet-101-PPM and ResNet-101-ASPP, which achieve performance of 75.65\%, 77.88\% and 78.61\% mIoU respectively. By stacking a DGCW module upon the base network, the performance is improved to 77.81\% ($\uparrow$2.16), 79.60\% ($\uparrow$1.72) and 80.43\% ($\uparrow$1.82) for ResNet-101, ResNet-101-PPM and ResNet-101-ASPP, respectively. DGCW module first weights each channel of each pixel's feature vector guided by intra-pixel channel-wise distance, then feeds the new feature map into two convolution layers for feature reconstruction. We further remove the process of feature vector channel weighting and directly feed the feature map into convolution layers. This modification decreases the mIoU to 75.90\% ($\downarrow$1.91\%), 78.19\% ($\downarrow$1.41\%) and 78.75\% ($\downarrow$1.68\%), respectively. It demonstrates that channel-wise weighting does play a crucial role in improving the performance. 

As described in Equation.\ref{get_weight}, a normalization layer is employed to transfer the channel-wise distance to channel-wise weight values. We simply divide each channel by the sum of all channel values of the feature vector as the normalization layer, denoted as \textbf{DBS}. Besides \textbf{DBS}, we also try another two normalization functions, including \textbf{softmax} and \textbf{tanh}. Compared to \textbf{DBS}, utilizing \textbf{softmax} decreases the mIoU to 77.46\% ($\downarrow$0.35\%), 79.19\% ($\downarrow$0.41\%), 79.95\% ($\downarrow$0.48\%) and utilizing \textbf{tanh} decrease the mIoU to 77.44\% ($\downarrow$0.37\%), 79.14\% ($\downarrow$0.46\%), 79.92\% ($\downarrow$0.51\%), respectively. Employing \textbf{DBS} yields the best performance. Thus, we use \textbf{DBS} as default in the following experiments. \\
\begin{figure}[t]
    \centering
    \includegraphics[width=3in]{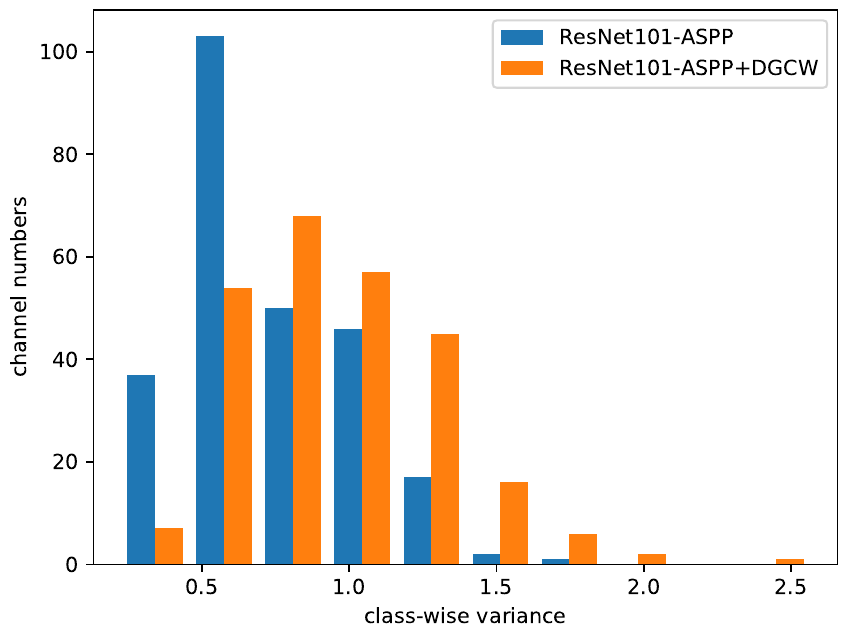}
    \caption{Counting of the channel numbers on different class-wise variance intervals.Results on ResNet101-ASPP and ResNet101-ASPP+DGCW are reported.}
    \label{discrime}
\end{figure}
\begin{table}[t]
    \centering
    \begin{tabular}{l c}
        \toprule
        Method & Mean IoU(\%)\\
        \midrule
        Baseline &  80.43\\
        Baseline + OHEM & 80.86\\
        Baseline + OHEM + MS & 81.53\\
        Baseline + OHEM + MS + Flip & 81.80\\
        \bottomrule
    \end{tabular}
    \caption{Ablation studies of some training and evaluating strategies on Cityscapes validation set.}
    \label{strategies}
\end{table}
\begin{table*}[tp]
    \centering
    \setlength{\tabcolsep}{4pt}
     \footnotesize
    \begin{tabular}{l|c|c c c c c c c c c c c c c c c c c c c}
    \toprule[1pt] \\
        Methods & \rotatebox{90}{Mean Iou} & \rotatebox{90}{road} & \rotatebox{90}{sidewalk} & \rotatebox{90}{building} & \rotatebox{90}{wall} & 
        \rotatebox{90}{fence} & 
        \rotatebox{90}{pole} & \rotatebox{90}{traffic light} & \rotatebox{90}{traffic sign} & \rotatebox{90}{vegetation} & \rotatebox{90}{terrain} & \rotatebox{90}{sky} & \rotatebox{90}{person} & \rotatebox{90}{rider} & \rotatebox{90}{car} & \rotatebox{90}{truck} & \rotatebox{90}{bus} & \rotatebox{90}{train} & \rotatebox{90}{motorcycle} & \rotatebox{90}{bicycle}\\
        \hline
        \hline
        PSPNet \dag\cite{pspnet} &78.4& & - & - & - & - & - & - & - & - & - & - & - & - & - & - & - & - & - & \\
        PSANet \dag\cite{psanet} & 78.6 & - & - & - & - & - & - & - & - & - & - & - & - & - & - & - & - & - & - & -\\
        \hline
        DGCWNet \dag& 80.8 & 98.7 & 87.0 & 93.7 & 58.5 & 63.3 & 70.3 & 77.8 & 81.2 & 93.9 & 73.4 & 95.7 & 87.8 & 73.8 & 96.1 & 71.6 & 83.2 & 78.2 & 71.3 & 79.0 \\
        \hline
        \hline
        Deeplab-v2 \ddag\cite{deeplabv2}& 70.4 & 97.9 & 81.3 & 90.3 & 48.8 & 47.4 & 49.6 & 57.9 & 67.3 & 91.9 & 69.4 & 94.2 & 79.8 & 59.8 & 93.7 & 56.5 & 67.5 & 57.5 & 57.7 & 68.8\\
        RefineNet \ddag\cite{refinenet}& 73.6 & 98.2 & 83.3 & 91.3 & 47.8 & 50.4 & 56.1 & 66.9 & 71.3 & 92.3 & 70.3 & 94.8 & 80.9 & 63.3 & 94.5 & 64.6 & 76.1 & 64.3 & 62.2 & 70.0\\
        DUC \ddag\cite{duc}& 77.6 & 98.5 & 85.5 & 92.8 & 58.6 & 55.5 & 65.0 & 73.5 & 77.9 & 93.3 & 72.0 & 95.2 & 84.8 & 68.5 & 95.4 & 70.9 & 78.8 & 68.7 & 65.9 & 73.8\\
        ResNet-38 \ddag\cite{resnet38}& 78.4 & 98.5 & 85.7 & 93.1 & 55.5 & 59.1 & 67.1 & 74.8 & 78.7 & 93.7 & 72.6 & 95.5 & 86.6 & 69.2 & 95.7 & 64.5 & 78.8 & 74.1 & 69.0 & 76.7\\
        PSANet \ddag\cite{psanet}& 80.1 & - & - & - & - & - & - & - & - & - & - & - & - & - & - & - & - & - & - & - \\ 
        Dense-ASPP \ddag\cite{denseaspp}& 80.6 & 98.7 & 87.1 & 93.4 & 60.7 & 62.7 & 65.6 & 74.6 & 78.5 & 93.6 & 72.5 & 95.4 & 86.2 & 71.9 & 96.0 & 78.0 & 90.3 & 80.7 & 69.7 & 76.8\\
        SPGNet \ddag\cite{spgnet}& 81.1 & 98.8 & 87.6 & 93.8 & 56.5 & 61.9 & 71.9 & 80.0 & 82.1 & 94.1 & 73.5 & 96.1 & 88.7 & 74.9 & 96.5 & 67.3 & 84.8 & 81.8 & 71.1 & 79.4\\
        CCNet \ddag\cite{ccnet}& 81.4 & - & - & - & - & - & - & - & - & - & - & - & - & - & - & - & - & - & - & -\\
        BFP \ddag\cite{bfp}& 81.4 & 98.7 & 87.0 & 93.5 & 59.8 & 63.4 & 68.9 & 76.8 & 80.9 & 93.7 & 72.8 & 95.5 & 87.0 & 72.1 & 96.0 & 77.6 & 89.0 & 86.9 & 69.2 & 77.6\\
        DANet \ddag\cite{danet}& 81.5 & 98.6 & 86.1 & 93.5 & 56.1 & 63.3 & 69.7 & 77.3 & 81.3 & 93.9 & 72.9 & 95.7 & 87.3 & 72.9 & 96.2 & 76.8 & 89.4 & 86.5 & 72.2 & 78.2\\
        \hline
        DGCWNet \ddag& 81.6 & 98.7 & 86.9 & 93.8 & 62.8 & 64.2 & 70.3 & 78.2 & 81.3 & 93.9 & 72.7 & 95.8 & 87.9 & 74.4 & 96.2 & 72.6 & 88.0 & 81.6 & 71.4 & 78.8\\
        \bottomrule
     \end{tabular}
     \caption{Performance comparison on Cityscapes test set. \dag denotes training with only training set; \ddag denotes training with both training set and validation set.}
    \label{city}
\end{table*}
\noindent \textbf{Comparison with other global context extraction methods.}
We further compare proposed DGCW module with several different global context extraction methods on Cityscapes validation set with ResNet-101, ResNet-101-PPM and ResNet-101-ASPP as base networks. Specifically, global context extraction methods utilized for comparison include: 1) \textbf{GAP} applies an image-level global average pooling on the output of base network, feeds the global feature into a $1\times 1$ convolution layer, then bilinearly upsamples it to the original size; 2) \textbf{SE} first performs a global average pooling on the input feature, then feeds the global features into two fully connected layers followed by a sigmoid function to get a channel-wise weight vector, finally multiplies the vector and input feature map. 3) \textbf{NL} updates each pixel's feature by weighting and summing feature vectors of all pixels, where weight values are obtained from the inter-pixel similarity. For a fair comparison, we set the output channel numbers of all convolution layers in these methods to 128. Note that the original \textbf{NL} does not downsample the feature map before processing it, while our DGCW module downsamples it to $\frac{1}{4}$ of the input size. For a fair comparison, we test both \textbf{NL} holding the input feature map's size and \textbf{NL} downsmapling it for 4 times, denoted as \textbf{NL-H} and \textbf{NL-D} respectively.

Experimental results are presented in Tab.~\ref{ablation2}. It can be observed that compared with base networks, adding \textbf{GAP} can slightly improve the performance, which verifies the effectiveness of capturing global contextual information. Adding \textbf{SE} achieves higher mIoU than adding \textbf{GAP} because it further models the channel interdependencies. By extracting the pixel-wise relationship, adding \textbf{NL-H} and \textbf{NL-D} outperforms adding \textbf{GAP} and \textbf{SE}. Our DGCW module achieves better performance than all compared global context extraction methods on all three base networks.\\

\textbf{Enhancement of feature discriminativeness. }By emphasizing the high-distance channels when modeling the pixel-wise relationship, DGCW module is able to increase the inter-class distance, thus enhancing feature’s discriminativeness. We conduct experiments to verify this point. We employ class-wise variance to measure the feature discriminativeness. Specifically, we first calculate the average feature vector of all pixels that belong to each class from all images in the dataset, denoted as class average feature. The class average feature $\mathbf{A}_{i}$ of class $i$ can be formulated as:
\begin{equation}
    \mathbf{A}_{i} = \frac{\sum_{j=1}^{N}\sum_{k=1}^{HW}\mathds{1}[y_{j}^{k}=i].\mathbf{F}_{j}^{k}}{\sum_{j=1}^{N}\sum_{k=1}^{HW}\mathds{1}[y_{j}^{k}=i]}.
\end{equation}
where $N$ is the number of images in the dataset, $H$ and $W$ are the height and width of feature map $\mathbf{F}$. $\mathbf{F}_{j}^{k}$ refers to feature vector of the $k$'th pixel from $j$'th image and $\mathds{1}[y_{j}^{k}=i]$ denotes a binary indicator denoting whether this pixel belongs to class $i$. After that, in each channel of the class average feature vector, we calculate the variance among all classes, getting the class-wise variance vector $\mathbf{V}\in \mathbb{R}^{D}$, where $D$ is the channel number of each class average vector. Channels with higher class-wise variance have higher inter-class distance, thus representing more discriminative properties. We calculate the class-wise variance on Cityscapes validation set utilizing feature maps from the last layer of deep networks, then count the channel numbers among different class-wise variance intervals. We conduct this experiment on ResNet101-ASPP and ResNet101-ASPP+DGCW, and employ a histogram to compare their results, which is shown in Fig.~\ref{discrime}.  It can be seen that after using DGCW module, the variance distribution of all channels moves to a larger direction. It demonstrates that DGCW module improves the feature discriminativeness effectively. 
\\
\begin{table}[t]
    \centering
    \begin{tabular}{l c c}
    \toprule
    Method & Backbone & Mean IoU(\%)\\
    \midrule
    FCN-8S \cite{fcn} & & 37.8\\
    CRF-RNN \cite{crfrnn} & & 39.3\\
    ParseNet \cite{parsenet} & & 40.4 \\
    BoxSup \cite{boxsup}& & 40.5 \\
    HO\_CRF \cite{hocrf}& & 41.3 \\
    PieceWise \cite{picewise}& & 43.3 \\
    VeryDeep \cite{verydeep}& & 44.5 \\
    DeepLab-v2 \cite{deeplabv2}& ResNet101-COCO & 45.7 \\
    RefineNet \cite{refinenet}& ResNet152 & 47.3 \\
    MSCI \cite{msci}& ResNet152 & 50.3 \\
    EncNet \cite{encnet}& ResNet101 & 51.7\\
    DANet \cite{danet}& ResNet101 & 52.6\\
    ANN \cite{ann}& ResNet101 & 52.8\\
    \midrule
    DGCWNet & ResNet101 & 53.9\\
    \bottomrule
    \end{tabular}
    \caption{Performance comparison on Pascal Context validation set }
    \label{context}
\end{table}
\noindent \textbf{Ablation of improving strategies.} We conduct some experiments to verify the effect of some widely used improving strategies, including online hard example mining (OHEM), multi-scale inputs for testing (MS) and random left-right flippping (Flip). All experiments in this part are based on ResNet-101-ASPP + DGCW. Experimental results are reported in Tab.~\ref{strategies}.
\begin{itemize}
    \item \textbf{OHEM:} We employ OHEM to handle the problem of class imbalance. The hard examples refer to pixels whose probabilities of the corresponding correct classes are smaller than $\theta$. We keep at least $K$ pixels within each batch in the training process. Following \cite{ocnet}, we set $\theta$ to 0.7 and $K$ to 100000. Utilizing OHEM improves performance to 80.86\%. 
    \item \textbf{MS:} Following \cite{acfnet}, we further adopt the multi-scale inputs strategies for testing. Specifically, we resize each test image with different scales [0.75, 1, 1.25, 1.5], feed these multi-scale images into network for prediction, then sum all prediction maps for final pixel-wise classification. This strategy improves the mIoU to 81.53\%.  
    \item \textbf{Flip} We further improve the performance by employing the left-right flipping strategy, gaining the highest 81.80\% mIoU on Cityscapes validation set.
\end{itemize}
\begin{figure*}[t]
    \centering
    \includegraphics[width=6.5in]{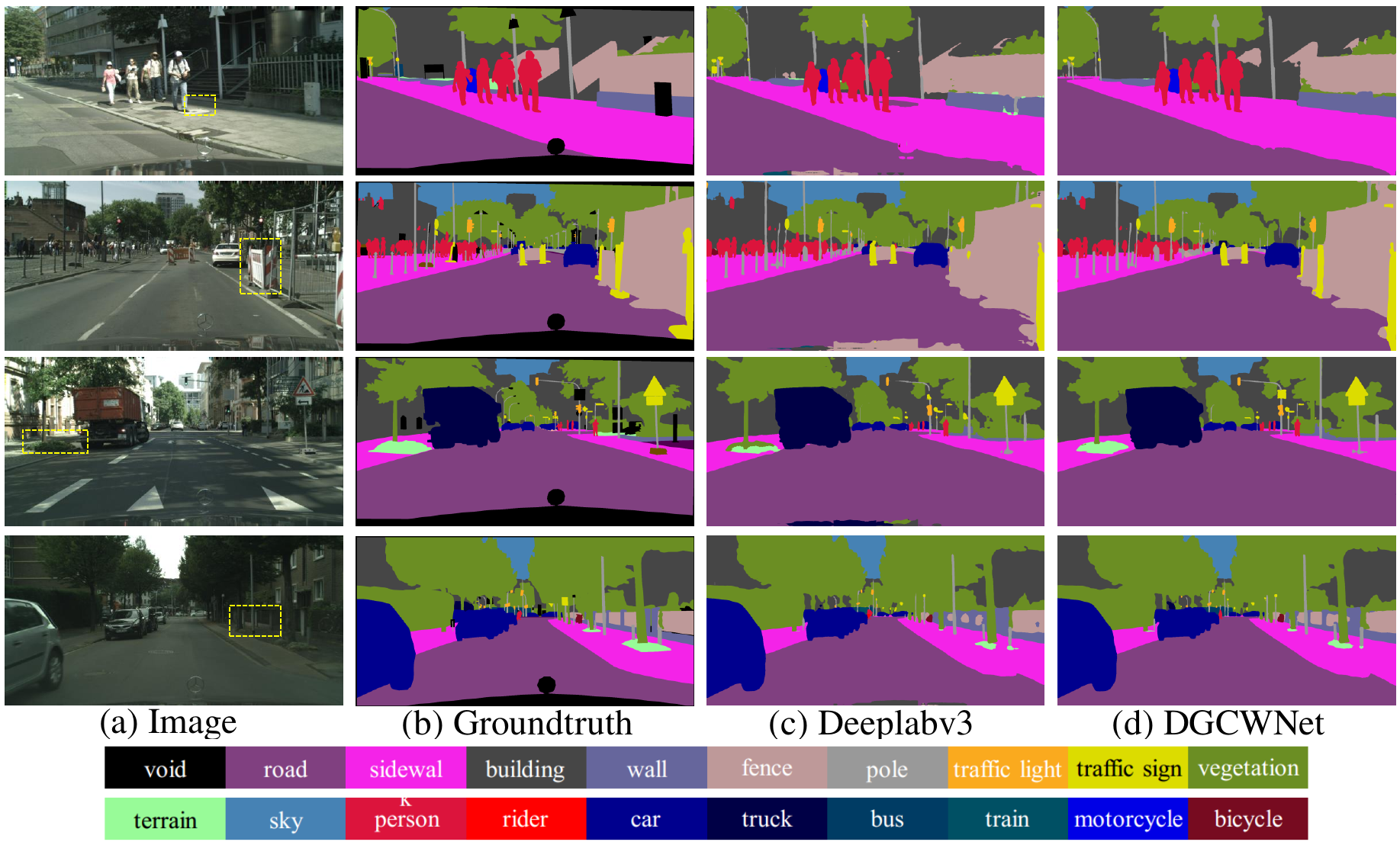}
    \caption{Visualization comparison on Cityscapes validation set.}
    \label{visual}
\end{figure*}
\subsubsection{Comparison with state-of-the-arts} 
We further compare DGCWNet with other state-of-the-art methods on Cityscapes test set and result is illustrated in Tab.~\ref{city}. The evaluation result is obtained by submitting prediction results to official evaluation severe. We use ResNet-101-ASPP + DGCW for comparison, as it gains the best performance among all tested DGCWNets. We train DGCWNet utilizing OHEM and test it using MS and Flip strategies. Following \cite{acfnet}, we report the comparison results of two cases, including using only training set and using both training set and validation set for training. In both cases, DGCWNet obtains excellent results. Even when only using training data, DGCWNet also outperforms PSANet and Dense ASPP that use both training data and validation data. After utilizing validation data further, DGCWNet gets the new state-of-the-art performance with an 81.6\% mIoU. Among the compared methods, many approaches extract the global pixel-wise context feature to improve performance. By capturing long-range pixel-wise dependencies from a new perspective, DGCWNet achieves better performance. 

Meanwhile, we provide the qualitative comparison between DGCWNet and Deeplabv3 in Fig.~\ref{visual}. We use yellow boxes to mark the challenging regions in images. Using DGCW module significantly improves the segmentation accuracy in these regions. 

\subsection{Experiments on PASCAL Context}
In this subsection, we conduct experiments on PASCAL Context dataset to further validate the effectiveness of DGCWNet. We also use the ResNet-101-ASPP + DGCW, and employ the same augmentation settings, training settings and testing settings as experiments on Cityscapes. The comparison results with other state-of-the-art methods are reported in Fig.~\ref{context}. DGCWNet yields an mIoU of 53.9\%, outperforming DANet and ANN that use the same backbone as ours. Moreover, DGCWNet even surpass some methods with deeper backbone \cite{refinenet, msci} or utilizing extra MS COCO dataset for pre-training \cite{deeplabv2}. 

\begin{table}[t]
    \centering
    \begin{tabular}{l c c}
    \toprule
    Method & Backbone & Mean IoU(\%)\\
    \midrule
    RefineNet \cite{refinenet} & ResNet152 & 40.70\\
    DSSPN \cite{dsspn}& ResNet101 & 43.68\\
    PSANet \cite{psanet} & ResNet101 & 43.77\\
    EncNet \cite{encnet} & ResNet101 & 44.68\\
    PSPNet \cite{pspnet} & ResNet101 & 43.29\\
    PSPNet \cite{pspnet} & ResNet269 & 44.94\\
    CCNet \cite{ccnet} & ResNet101 & 45.22\\
    ANN \cite{ann} & ResNet101 & 45.24\\
    \midrule
    DGCWNet & ResNet101 & 45.51\\
    \bottomrule
    \end{tabular}
    \caption{Performance comparison on ADE20K validation set }
    \label{ade}
\end{table}

\subsection{Experiments on ADE20K}
We further conduct experiments on ADE20k, and evaluate our methods on its validation set. The comparison results with other methods are shown in Tab.~\ref{ade}. DGCWNet achieves highest performance (45.51\% mIoU) among all compared methods.

\section{Conclusion}
In this paper, we propose a Distance Guided Channel Weighting Network (DGCWNet) for semantic segmentation. By fully utilizing the discriminative features while ignoring the useless features, DGCWNet enhance features' representiveness. Moreover, by modeling the pixel-wise correlation, it is also able to capture the long-range dependencies. Ablation studies and performance comparison on several popular datasets demonstrate the effectiveness of proposed methods.

\section{Acknowledgment}
This work was supported by the National Natural Science Foundation of China (NSFC) under grant No. 61375025, No. 61075011 and No. 60675018, and the Scientific Research Foundation for the Returned Overseas Chinese Scholars from the State Education Ministry of China.

{\small
\bibliographystyle{ieee_fullname}
\bibliography{egbib}
}

\end{document}